\documentclass[letterpaper, 10 pt, conference]{ieeeconf}
\IEEEoverridecommandlockouts                         
\usepackage{cite}
\usepackage[ruled,linesnumbered]{algorithm2e}
\usepackage[normalem]{ulem}
\usepackage{color}
\let\labelindent\relax
\usepackage{enumitem}

\usepackage[acronym,nomain]{glossaries}
\usepackage{amsmath,amssymb}
\usepackage{bm}
\usepackage{upgreek}
\usepackage{url}

\usepackage{xcolor} 
\usepackage{tikz}
\usetikzlibrary{fit,calc}
\newcommand*{\tikzmk}[1]{\tikz[remember picture,overlay,] \node (#1) {};\ignorespaces}
\newcommand{\boxit}[1]{\tikz[remember picture,overlay]{\node[yshift=2pt,fill=#1,opacity=.25,fit={($(A)+(.075,.15)$)($(B)+(.175\linewidth,.8\baselineskip)$)}] {};}\ignorespaces}
\newcommand{\boxittwo}[1]{\tikz[remember picture,overlay]{\node[yshift=3pt,fill=#1,opacity=.25,fit={(A)($(B)+(.775\linewidth,.8\baselineskip)$)}] {};}\ignorespaces}
\newcommand{\boxitthree}[1]{\tikz[remember picture,overlay]{\node[yshift=3pt,fill=#1,opacity=.25,fit={(A)($(B)+(.715\linewidth,.8\baselineskip)$)}] {};}\ignorespaces}
\colorlet{pink}{red!40}
\colorlet{blue}{cyan!60}

\newglossary[slg]{symbolslist}{syi}{syg}{Symbolslist}
\makeglossaries                                   %

\newglossaryentry{psistl}{
    name=\ensuremath{\Psi},
    description={Event-based STL Specification}}

\newglossaryentry{trueprop}{
    name=\ensuremath{\sigma_{t}},
    description={Set of propositions that are \ensuremath{True} at time \ensuremath{t}}}

\newglossaryentry{robots}{
    name=\ensuremath{n},
    description={number of robots}}

\newglossaryentry{state}{
    name=\ensuremath{\textbf{X}_t},
    description={State of the system and dynamic obstacles at time \ensuremath{t}}}
    
\newglossaryentry{time}{
    name=\ensuremath{t},
    description={time}}
    
\newglossaryentry{psinew}{
    name=\ensuremath{\Psi_{new}},
    description={Modified Event-based STL Specification}}
\newglossaryentry{opset}{
    name=\ensuremath{\textbf{op}},
    description={set of method of adding new specifications }}
  \newglossaryentry{op}{
    name=\ensuremath{op},
    description={method of adding new specification ($\wedge$ or $\vee$)}}

\newglossaryentry{psistl'}{
    name=\ensuremath{\Psi'},
    description={Event-based STL Specification added during execution via conjunction or disjunction}}
    
\newglossaryentry{pimuset}{
    name=\ensuremath{\bm{\pi}_{\mu,[a,b]\rightarrow\mu',[a',b']}},
    description={list of modified proposition in \gls{psinew}}}
    
\newglossaryentry{pimu}{
    name=\ensuremath{\pi_{\mu,[a,b]\rightarrow\mu',[a',b']}},
    description={A modified proposition in \gls{psinew}}}
    
\newglossaryentry{piab}{
    name=\ensuremath{\pi_{[a,b]\rightarrow[a',b']}},
    description={A modified proposition in \gls{psinew}}}
    
\newglossaryentry{psistlset}{
    name=\ensuremath{\bm{\Psi}'},
    description={list of modifications \gls{psistl'} added during execution and the method of addition \gls{delta'}}}
    
\newglossaryentry{rhoset}{
    name=P,
    description={Set of activated propositions and their robustness}}
\newglossaryentry{psimu}{
    name=\ensuremath{\Psi_{new,\mu \rightarrow \mu'}},
    description={Modification of a predicate in an Event-based STL Specification}}

\newglossaryentry{psiab}{
    name=\ensuremath{\Psi_{new,[a,b] \rightarrow [a',b']}},
    description={Modification of the timing bounds in an Event-based STL Specification}}

\newglossaryentry{buchi}{
    name=\ensuremath{B},
    description={B\"{u}chi automaton}}
\newglossaryentry{buchiset}{
    name=\ensuremath{\mathcal{B}},
    description={Set of B\"{u}chi automata}}
    
\newglossaryentry{states}{
    name=\ensuremath{s_i},
    description={State \ensuremath{i} in B\"{u}chi automaton}}

\newglossaryentry{initialstate}{
    name=\ensuremath{s_0},
    description={Initial state in B\"{u}chi automaton}}
    
\newglossaryentry{potentialstates}{
    name=\ensuremath{s_{\mathcal{B}}},
    description={Set of current states for B\"{u}chi automata in \ensuremath{B}}}
    
\newglossaryentry{abstractedpred}{
    name=\ensuremath{\Pi_\mu},
    description={Set of abstracted predicates}}
    
\newglossaryentry{robustset}{
    name=P\ensuremath{(\Pi_{\bm{\upsigma},True})},
    description={robustness tuple}}

\newglossaryentry{predicatefunc}{
    name=\ensuremath{h(\textbf{X}_t)},
    description={predicate function}}
    
\newglossaryentry{setpredicatefunc}{
    name=\ensuremath{H},
    description={Set of predicate functions}}

\newglossaryentry{cbf}{
    name=\ensuremath{CBF},
    description={Set of control barrier functions for predicates in \gls{abstractedpred}}}
    
\newglossaryentry{activated}{
    name=\ensuremath{\Pi_{\mu_{act}}},
    description={Set of abstracted predicates to be activated}}

\newglossaryentry{control}{
    name=\ensuremath{u},
    description={Control for robot}}
    
\newglossaryentry{prefailure}{
    name=\ensuremath{preF},
    description={Generated pre-failure warnings}}
    
\newglossaryentry{trans}{
    name=\ensuremath{trans^*},
    description={Set representing the best transition from all B\"{u}chi automatons}}
    
\newglossarystyle{notationlong}{%
\setglossarystyle{long}
  {\end{longtable}}%

}
\title{\LARGE \bf
Online Modifications for Event-based Signal Temporal Logic Specifications 
}

\author{David Gundana and Hadas Kress-Gazit
\thanks{D. Gundana and H. Kress-Gazit are with Sibley School of Mechanical and Aerospace Engineering, Cornell University, Ithaca, NY, 14853 USA. e-mail: \{dog4,hadaskg\}@cornell.edu. This work is supported by the National GEM Consortium, Cornell Sloan Fellowship, and NSF IIS-1830471 
}%
}

\begin{document}

\maketitle
\thispagestyle{empty}
\pagestyle{empty}


\begin{abstract}
In this paper we present a grammar and control synthesis framework for online modification of Event-based Signal Temporal Logic (STL) specifications, during execution. These modifications allow a user to change the robots' task in response to potential future violations, changes to the environment, or user-defined task design changes. In cases where a modification is not possible, we provide feedback to the user and suggest alternative modifications. We demonstrate our task modification process using a Hello Robot Stretch satisfying an Event-based STL specification. 

\end{abstract}

\section{INTRODUCTION}

High-level specifications can be used to describe complex robotic tasks, and there exist algorithms to synthesize control from them~\cite{Loizou2004,Filippidis2012,Kloetzer2007,Chen2018,Kress-Gazit2018,kress2009temporal,raman2014model}. These tasks may include time critical objectives, response to environment events or disturbances, and complex sequencing of actions. To describe these high-level specifications, researchers have used temporal logics such as Linear Temporal Logic (LTL) \cite{clarke2018model} for discrete abstractions of systems and Signal Temporal Logic (STL) \cite{Maler2004} for behaviors defined over continuous signals. A crucial aspect of using such formalisms in robotics in the ability to automatically synthesize robot control from specifications~\cite{Kress-Gazit2018}.

Approaches to control synthesis for STL specifications include Model Predictive Control (MPC), Control Barrier Functions (CBFs), or other Mixed-integer Linear Programs (MILP); MPC methods satisfy specifications by converting them, and the dynamics of the system, into a receding-horizon optimization problem ~\cite{raman2014model,raman2015reactive,farahani2015robust,charitidou2021barrier}. These methods are useful in that they provide guarantees on the satisfaction of a specification over the prediction horizon, but they do not scale to long time horizons or complex specifications, due to the increase in decision variables. Other MILP solutions to synthesizing control satisfying STL specifications \cite{buyukkocak2021planning, sun2022multi} attempt to improve the computational efficiency by decomposing the specification into subtasks or waypoints. Work from \cite{kurtz2022mixed} provide methods to satisfy STL specifications using Mixed-integer Convex Programming while reducing the number of binary decision variables in the optimization problem. While these methods maintain guarantees on the satisfaction of the specification, they still rely on solving optimization problems whose decision variables increase with the complexity of a specification. This complexity makes specifications difficult to satisfy in real time and in the presence of uncontrolled environment events. Work from \cite{lindemann2018control,lindemann2019robust,lindemann2019decentralized, buyukkocak2022control} introduced the use of time-varying CBFs to satisfy STL specifications. Time-varying CBFs improves the computational efficiency of synthesizing control satisfying specifications as there are less decision variables in the optimization problem used to generate control at each time step. These techniques are computationally efficient enough to create control for complex specifications in real time; however, the cannot provide the guarantees, as in the MPC approach. 

In prior work~\cite{gundana,gundana2022event}, we have introduced Event-based STL to describe tasks that include reactivity to external environment events. To do this, we defined a grammar and a control synthesis approach that is computationally efficient enough to enable the system to respond in real-time to events, while also satisfying timing constraints. To control the system, we compose, using an automata based framework, time-varying CBFs \cite{Lindemann1} such that the system either completes the task, or provides pre-failure warnings. 

In this paper, we provide the ability to update the specification, i.e. the task, \textit{while the robots are executing it}. During execution, the environment may change causing previously satisfiable specifications to become unsatisfiable. For example, if operating in a space with humans, while avoiding collisions with the human, a robot may require more time to reach a goal. In our previous work, while we would provide a pre-failure warning, this specification would be violated as the time constraints from the Event-based STL specification were specified before the execution. Additionally, there may exist new tasks that the operator would like the robots to accomplish during execution, in addition to the original task. For example, while a robot is patrolling two rooms, the user may want to add a third room for the robot to patrol. In these cases, instead of stopping, synthesizing a new specification, and restarting the execution, the user will add the new requirements while continuing to satisfy the original task and its timing constraints. 

We present a grammar for modifying Event-based STL specification (Sec.~\ref{sec:modificationGrammar}) and a framework for online control synthesis (Sec.~\ref{sec:synthesis}) that seamlessly incorporates the modified specifications into the robot control. The modified specification is satisfied, if possible, given the robots bounded control input. When a satisfying controller is not found, we provide automatically generated feedback to the user, enabling them to refine the modifications. In modifying specifications, we allow changing the parameters of the currently executing specification, and adding new tasks to the existing specification. 

There has been previous work modifying tasks in real time. Work from \cite{ding2005logic,guo2015multi} show how tasks are updated based on the addition of new information from sensors and knowledge base (real-time plan configuration). In \cite{guo2015multi} specifications are divided into hard and soft sub-specifications and plans to satisfy the specification are generated before execution begins.
Hard sub-specifications such as safety constraints must always be satisfied while least-violating solutions are found which make progress towards the satisfaction of soft sub-specifications. During execution, the plan is updated and new solutions are found based on environment knowledge updates from the operating robots. In our work, we allow the user to directly update a task based on their preference or changes in goals. 

Using existing methods such as MPC ~\cite{raman2014model,raman2015reactive,farahani2015robust,charitidou2021barrier} or MILP solutions \cite{buyukkocak2021planning, sun2022multi}, one can update the specification; however, the new specification would initiate re-synthesis and, as stated above, this is a computationally inefficient. In our work, we provide an efficient solution to modify tasks online without the need to stop or pause execution, in most cases; we discuss computation limitations that require execution to pause for large specification modifications.

\textbf{Assumptions}: We assume that all robots in the system have knowledge of the state of all other robots in the system, static obstacles, and dynamic obstacles. However, the robots do not know or try to reason about the behavior of other robots and moving obstacles. Additionally, we assume that the specification is not trivially violated at initialization. 

\textbf{Contributions:} In this paper we present three main contributions: 1) a grammar for modifying Event-based STL specifications that formally defines how specifications can be modified during execution, 2) an automated control synthesis framework for satisfying modified Event-based STL specifications at run time, 3) automatically generated feedback that notifies an operator when a modification may not have the desired affect or when a specification may be violated in the future due to the bounded control of the robots. Our proposed framework allows an operator to change the desired behavior of a task during execution if a new behavior is required or if the original task may become infeasible. We demonstrate our framework using a Hello Robot Stretch. 

\section{PRELIMINARIES}
\subsection{System and Environment Model}
The state and bounded control input of the system at time $t$ are denoted by $\textbf{x}_t \in \mathbb{R}^n$ and $\textbf{u}_t \in \textbf{U} \subset \mathbb{R}^m$, respectively. Eqn. (\ref{dynamics}) describes the discrete time dynamical system we control. 
\begin{equation}
    \textbf{x}_{t+1} = f(\textbf{x}_t) + g(\textbf{x}_t)\textbf{u}_t,
    \label{dynamics}
\end{equation}

where the functions $f: \mathbb{R}^n \rightarrow \mathbb{R}^n$ and $g: \mathbb{R}^n \rightarrow \mathbb{R}^{n \times m}$ are locally Lipschitz continuous. 

The environment in which the system is operating contains static and dynamic obstacles; static obstacles, such as walls and other objects whose positions do not change during execution, are known a priori. The position of dynamic obstacles such as humans, robots, and other objects, $\textbf{x}_{dyn,t}$ are known to the system during execution. An Event-based STL specification may require that the system avoid these dynamic obstacles (collision-avoidance) or operate with them. For example, a specification may require a robot to navigate to and grasp a moving object. For convenience we denote the state of the system and dynamic obstacles as $\gls{state} = (\textbf{x}_t, \textbf{x}_{dyn,t})$.

\subsection{LTL and B\"{u}chi Automata}
Linear Temporal Logic~\cite{clarke2018model} has been used for synthesis in robotic systems~\cite{Kress-Gazit2018}. An LTL formula is constructed from Boolean propositions $\pi \in AP$ where $AP$ is a set of atomic propositions. The grammar is as follows:  
\begin{equation}
    \gamma ::= \pi |\  \neg \gamma \ |\  \gamma_1 \vee \gamma_2 \ |\  X \gamma\ |\ \gamma_1 U \gamma_2.
\end{equation}
Where $\neg$ and $\vee$ are the Boolean operators ``not" and ``or" respectively and $X$ and $U$ are the temporal operators ``next" and ``until" respectively. The semantics of LTL are defined over an infinite sequence $\sigma = \sigma_1,\sigma_2,\ldots$ of truth assignments to the Boolean propositions $\pi \in AP$ where $\sigma_i$ represents the set of $AP$ that are $True$ at position $i$. The full Semantics of LTL as well as the definitions for other Boolean and temporal operators can be found in \cite{clarke2018model}.


A B\"{u}chi automata is a tuple $B = (S,s_0,\Sigma,\delta,F)$. $S$ is the set of states, where $s_0 \in S$ is the initial state and $F \subseteq S$ is a set of accepting states, $\Sigma$ is a finite input alphabet, and $\delta : S \times \Sigma \rightarrow 2^S$ is a transition function. A transition occurs between states when the label on the transition between the two states is evaluated to $True$. A run of a B\"{u}chi automata, $B$, on an infinite word $w = w_1,w_2,w_3,\ldots, w_i\in2^{\Sigma}$ is an infinite sequence of states $s_1,s_2,s_3,\ldots$, s.t. $\forall j \geq 1, (s_{j-1},w_j,s_j) \in \delta$. A run of $B$ is accepting if and only if $\inf(\omega)\cap F \neq \emptyset$, where $\inf(\omega)$ represents a set of states that are visited infinitely often on the input word $w$. Given an LTL formula $\gamma$, we can construct a B\"{u}chi automaton $B_\gamma$ that accepts infinite words if and only if they satisfy $\gamma$. In the following we use \cite{Duret-Lutz2016} for this construction.

\subsection{B\"{u}chi Intersection}
\label{sec:buchiIntersect}
Given two B\"{u}chi automata, $B_{\gamma_1} = (S^1,s_0^1,\Sigma,\delta^1,F^1)$ and $B_{\gamma_2} = (S^2,s_0^2,\Sigma,\delta^2,F^2)$, their intersection $B_{\gamma_1} \times B_{\gamma_2}$ is a B\"{u}chi automaton whose accepting traces satisfy $\gamma_1 \wedge \gamma_2$. An accepting run of this B\"{u}chi intersection must visit the accepting states of both B\"{u}chi automata infinitely often. From \cite{clarke2018model}: $B_\gamma = B_{\gamma_1} \times B_{\gamma_2} = \{S^1 \times S^2 \times \{0,1,2\}, s_0^1 \times s_0^2 \times \{0\}, \Sigma,\Delta, S^1 \times S^2 \times \{2\}\}$. A transition in the resultant B\"{u}chi automata is defined as $\Delta = ((g,q,i),\alpha,(g',q',j))$. A transition exists in  $\Delta$ if the transition $(g,\alpha,g') \in \delta^1$ and $(q,\alpha,q') \in \delta^2$, where $\alpha \in \Sigma$. 
The accepting states of $B_\gamma$ ensure that accepting states in $B_{\gamma_1}$ and $B_{\gamma_2}$ are visited infinitely often. Further details can be found in~\cite{clarke2018model}. 

In this work we follow the B\"{u}chi intersection definition from \cite{fang2022automated} which does not assume both B\"{u}chi automata share the same finite alphabet $\Sigma$. Given two B\"{u}chi automata, $B_{\gamma_1}$ and $B_{\gamma_2}$ with input alphabets $\Sigma_1$ and $\Sigma_2$, we define the input alphabet of the resultant intersection to be $\Sigma = \Sigma_1 \cup \Sigma_2$ and use the projection on the original alphabets when determining transitions.  

\subsection{Event-based STL}
Event-based STL \cite{gundana2022event} is defined over Boolean predicates $\mu$ representing the system state, and uncontrolled environment events which are represented as Boolean propositions $\pi$. The truth value of $\mu$ is determined by the evaluation of a predicate function \gls{predicatefunc} as follows: 
\begin{equation}
\mu ::= 
    \begin{cases}
        False & \Rightarrow h(\textbf{X}_t) < 0\\
        True & \Rightarrow h(\textbf{X}_t) \geq 0.\\
    \end{cases}  
    \label{predicate}
\end{equation}

An Event-based STL specification \gls{psistl} is constructed using the following syntax: 
\begin{align}
    \varphi ::=  &\mu \ |\ \neg \mu\ | \   \varphi_1 \wedge \varphi_2 \ | \   \varphi_1 \vee \varphi_2,\\
    \alpha ::= &\pi \  | \ \neg \alpha \ |\   \alpha_1 \wedge \alpha_2, \  \\
    \begin{split}
    \Psi ::= &G_{[a, b]}\ \varphi \ |\ F_{[a, b]}\ \varphi \ | \ \varphi_1 \ U_{[a,b]} \ \varphi_2 \ |\\ &\ G(\alpha \Rightarrow \ \Psi) \ | \  \Psi_1 \wedge \Psi_2 \  | \ \Psi_1 \vee \Psi_2,
    \end{split}
\end{align}
where $\varphi$ is a formula over predicates $\mu$, $\alpha$ is a Boolean formula over $\pi \in AP$, $\{a,b\} \in \mathbb{R}^+$ are timing bounds for a formula, $F$ is ``Eventually", $G$ is ``Always", U is ``Until", and $\Rightarrow$ is implication. The operator $G$ with no timing bound is assumed to have the timing bounds of $[0,\infty]$. The full definition of the semantics of Event-based STL is in \cite{gundana2022event}.

The approach in \cite{gundana2022event} synthesizes control for Event-based STL formulas $\Psi$ by first abstracting $\Psi$ into an LTL formula, creating the corresponding B\"{u}chi automata and using it to sequentially compose Control Barrier Functions (CBFs) corresponding to the predicates in $\Psi$. CBFs define safe-sets and ensure that they are forward invariant \cite{Wieland2007,Wang2017,Ames2017}. Using techniques from \cite{lindemann2018control, gundana} the approach creates templates that are composed and instantiated at runtime with appropriate timing bounds, in response to environmental events. 

\section{MODIFICATION GRAMMAR}
\label{sec:modificationGrammar}
The main contribution of this paper is to enable users to modify specification during execution. We specify the allowable modifications using the following grammar: 
\begin{equation}
\begin{split}
        \gls{psinew} ::&= \gls{psistl} \ | \ \gls{psinew} \vee  \gls{psistl'} \ |\  \gls{psinew} \wedge \gls{psistl'}\ |\\& \gls{psimu} \ |\  \gls{psiab}
\end{split}
\label{modify}
\end{equation}

This grammar describes four different types of modifications that we allow: 
\begin{enumerate}
    \item $\gls{psinew} \vee  \gls{psistl'}$ describes the addition of an Event-based STL specification using disjunction. Here, the system must satisfy either the original OR the new specification 
    \item $\gls{psinew} \wedge  \gls{psistl'}$ describes the addition of an Event-based STL specification $\gls{psistl'}$ using conjunction. Here the original specification and the new specification must both be satisfied
    \item \gls{psimu} modifies a predicate $\mu$ within a specification \gls{psistl}, i.e. replace $\mu$ with $\mu'$ 
    \item \gls{psiab} modifies timing bounds of an existing temporal operator in \gls{psistl}, i.e. replace $[a,b]$ with $[a',b']$. This modification changes the timing bounds of all predicates $\mu \in \varphi$ in a specific formula $F_{[a,b]}\varphi$, $G_{[a,b]}\varphi$, or $\varphi_1 U_{[a,b]}\varphi_2$.
\end{enumerate}

\section{PROBLEM STATEMENT}
Given an Event-based STL specification \gls{psistl}, user defined modifications \gls{psinew} given during execution, uncontrolled environment external events $\alpha$, and the state of the system and environment \gls{state}, find a control strategy \gls{control} to satisfy the modified Event-based STL specification during execution. If a control strategy is not found, provide feedback to the user. When a modification $\gls{psinew}$ is made to a specification \gls{psistl}, the original specification is replaced for the remainder of the execution; we allow multiple modifications during system execution. 



\section{CONTROL SYNTHESIS FOR ONLINE MODIFICATIONS}
\label{sec:synthesis}

\subsection{Synthesis Overview} 
In this paper, we follow the control synthesis approach defined in \cite{gundana2022event} to automatically generate control and feedback for an Event-based STL specification. We use Algo. \ref{algo:Control} to briefly outline this procedure and highlight the changes needed for online modification (lines highlighted in blue). First, we pre-process the specification by abstracting \gls{psistl} as an LTL formula $\gamma$ and generating a B\"{u}chi automaton \gls{buchi} with an initial state \gls{initialstate} (line \ref{line:prepare} of Algo. \ref{algo:Control}). While abstracting the specification, we record the continuous semantics of each predicate $\mu$ and temporal operator in the Event-based STL specification $\gls{psistl}$ through a set of abstracted propositions \gls{abstractedpred}. The propositions in \gls{abstractedpred} contain information on the timing bounds $[a,b]$ and predicate functions $h(\textbf{x}_t) \in \gls{setpredicatefunc}$ of each predicate in the specification that they abstract. To satisfy an Event-based STL specifications, we create time-varying control barrier function templates \gls{cbf} using the continuous values recorded in  each proposition in \gls{abstractedpred}. Using work from \cite{Lindemann1,gundana} these time-varying control barrier functions ensure an Event-based STL task is satisfied given the timing constraints for each predicate function $h(\textbf{x}_t) \in \gls{setpredicatefunc}$.

Following the pre-processing step and initialization, execution of the specification begins and, at each timestep, we choose an instantaneously-robust trace through \gls{buchi} to an accepting state that satisfies $\gamma$ given the set of propositions that are $True$ at time $t$ and the state of the system \gls{state} (line \ref{line:transitions} of Algo. \ref{algo:Control}). Instantaneous robustness is a measure that determines how robustly a specification is satisfied at a timestep in an execution given the current state of the environment and system. We determine the instantaneous robustness by evaluating the first transition of all accepting traces of the system. Within the first transition, we assign a robustness score to each proposition in a transition which measures how robustly the system can satisfy that proposition. We choose the transition (and trace) that contains the highest minimal robustness score of all possible transitions.
Further details on the procedure for satisfying an Event-based STL specification that maximizes instantaneous robustness are in \cite{gundana2022event}.

In the following sections we show how we build on this control synthesis framework to make online modifications to an executing Event-based STL specification. The main contributions of this paper, highlighted in blue in Algo. \ref{algo:Control}, are contained in the use of the newly introduced modification grammar to make modifications at runtime (lines \ref{line:neq}-\ref{line:endneq} in Algo. \ref{algo:Control}) and how transitions in $\gamma$ are chosen to satisfy the modified Event-based STL specification (line \ref{line:choose} in Algo. \ref{algo:Control}).  

\begin{algorithm}[h]
\SetAlgoLined
\SetKwInOut{Input}{Input}
\SetKwInOut{Output}{Output}
\SetKwProg{Initialization}{Initialization}{}{}
\Input{\gls{psistl}, \gls{trueprop}, \gls{state}, \gls{psinew}}
\Output{\gls{control},  \gls{prefailure}}
$[$\gls{abstractedpred}, \gls{buchi}, \gls{setpredicatefunc}, \gls{cbf}, \gls{initialstate}$] = prepSpec($\gls{psistl}$)$\;\label{line:prepare}

\tikzmk{A}
$\gls{buchiset} = [\gls{buchi}]$\;\label{line:buchiStates}
$\gls{potentialstates} =[\gls{initialstate}]$\;\label{line:potStates}
\tikzmk{B}
\boxit{blue}
\While{True}{
\tikzmk{A}
\If{$\gls{psistl} \neq \gls{psinew}$ \label{line:neq}}{
$[\gls{buchiset}, \gls{potentialstates}, \gls{abstractedpred}, \gls{cbf}, \gls{setpredicatefunc}, \gls{psistl}] = modify(\gls{buchiset},  \gls{potentialstates}, \gls{abstractedpred}, \gls{cbf}, \gls{setpredicatefunc}, \gls{psistl}, \gls{psinew})$\;\label{line:modify}
}\label{line:endneq}
\tikzmk{B}
\boxittwo{blue}
\gls{rhoset}$\  = [\ ]$\;
\For{$i = 1$ to $|\gls{buchiset}|$\label{line:for}}{
$[s, \gls{activated}, \gls{robustset}]\!= pickT(\gls{trueprop},\gls{state}, \gls{buchiset} [i], \gls{setpredicatefunc}, \gls{potentialstates} [i])$\; \label{line:transitions}

$\gls{potentialstates} [i] = s$\;

\gls{rhoset}$[i] =  (\gls{activated},\gls{robustset})$\;

}
\tikzmk{A}
$[\gls{activated}, \gls{buchiset}, \gls{potentialstates} ]= chooseProps($P$,\gls{buchiset}, \gls{potentialstates})$\;\label{line:choose}
\tikzmk{B}
\boxitthree{blue}
\gls{control}$= generateControl($\gls{activated}, \gls{time}, \gls{state},\gls{cbf}$)$\;\label{line:commands}

\gls{prefailure}$= preFailure($\gls{trueprop}, \gls{state}, \gls{buchiset}, \gls{setpredicatefunc}, \gls{potentialstates}$)$\; \label{line:pref}

}

\caption{Control Synthesis for Event-based STL}
\label{algo:Control}
\end{algorithm}

\noindent \textbf{Example:} We use the following example to illustrate the process of automatically generating control for modified specifications: 
\begin{equation}
    \gls{psistl}_{alarm} = G(alarm \Rightarrow F_{[0,7]}(\parallel \textbf{x}_t - [3,4]\parallel < 1))
\end{equation}

For a robot with state $\textbf{x}_t = [x_t,y_t]$, this specification states that if the event $alarm$ is sensed, then the distance from the robot and point $[3,4]$ must eventually, within 7 seconds, be less than 1.

\subsection{Online Modifications of Event-based STL Specifications}
During execution an operator may make modifications to a specification online according to the grammar defined in Sec. \ref{sec:modificationGrammar}. If the modified specification $\gls{psinew} \neq \gls{psistl}$  (lines \ref{line:neq}-\ref{line:endneq} in Algo. \ref{algo:Control}), the task execution is modified following Algo. \ref{algo:modification}: 

\begin{algorithm}[h]
\SetAlgoLined
\SetKwInOut{Input}{Input}
\SetKwInOut{Output}{Output}
\SetKwProg{Initialization}{Initialization}{}{}

\Input{\gls{buchiset}, \gls{potentialstates}, \gls{abstractedpred}, \gls{cbf}, \gls{setpredicatefunc}, \gls{psistl}, \gls{psinew}}
\Output{\gls{buchiset}, \gls{potentialstates}, \gls{abstractedpred}, \gls{cbf}, \gls{setpredicatefunc}, \gls{psistl}}

$[\gls{psistlset},\gls{opset},\gls{pimuset}] = findMods(\gls{psistl},\gls{psinew})$\;\label{line:findMods} 

\For{$i = 1 \ to\  |\gls{psistlset}|$}{
    $[\gls{abstractedpred}^{'}, \gls{buchi}^{'}, \gls{setpredicatefunc}^{'}, \gls{cbf}^{'}, \gls{initialstate}^{'} ] = prepSpec(\gls{psistlset} [i])$\;\label{line:prepMod}

    $\gls{abstractedpred} \leftarrow \gls{abstractedpred} \cup \  \{\gls{abstractedpred}^{'}\}$\;\label{line:addPred}
    $\gls{setpredicatefunc} \leftarrow \gls{setpredicatefunc} \cup \  \{\gls{setpredicatefunc}^{'}\}$\;\label{line:addsetpred}
    $\gls{cbf} \leftarrow \gls{cbf} \cup \  \{\gls{cbf}^{'}\}$\;\label{line:addcbf}
    
    \uIf{$\gls{opset} [i] = \wedge$}{
    $[\gls{buchiset}, \gls{potentialstates}] = intersectB(\gls{buchiset}, \gls{potentialstates}, \gls{buchi}^{'},\gls{initialstate}^{'})$\;\label{line:buchiIntersect}
    
    }

    \ElseIf{$\gls{op} [i] = \vee$}{
    $\gls{buchiset} = [\gls{buchiset},  \gls{buchi}^{'}]$\;\label{line:addbuchi}
    
    $\gls{potentialstates} = [\gls{potentialstates},  \gls{initialstate}^{'}]$\;\label{line:addstate}

    }

}

\For{$i = 1 \ to\  |\gls{pimuset}|$}{
$[\gls{abstractedpred}, \gls{cbf}, \gls{setpredicatefunc},\gls{psistl}] = modifyProp(\gls{abstractedpred}, \gls{cbf}, \gls{setpredicatefunc}, \gls{psistl}, \gls{pimuset} [i])$\;\label{line:modifyprop}
}

\gls{psistl} = \gls{psinew}\;\label{line:setequal}

\caption{$modify$ (Update Control)}
\label{algo:modification}

\end{algorithm}

\textbf{\textit{findMods}} \textbf{(line \ref{line:findMods})}: Given the original specification \gls{psistl} and the modified specification \gls{psinew}, we first parse the modifications. 
The list $\gls{psistlset} = [\gls{psistl}'_1, \gls{psistl}'_2,\ldots, \gls{psistl}'_i]$ contains the new formulas   $\gls{psistl}'_i$  that are added to the original specification $\gls{psistl}$; the list $\gls{opset} = [\gls{op}_1, \gls{op}_2, \ldots,\gls{op}_i]$ captures which Boolean operator is used, $\gls{op}_i \in \{\vee, \wedge\}$. 
Modifications to the timing bounds and predicate of a specification are captured by  $\gls{pimuset} = [{\gls{pimu}}_1, {\gls{pimu}}_2,\ldots, {\gls{pimu}}_i]$ where each element in the list represents a modification of the predicate or timing bounds of an existing proposition $\pi_\mu$ to a proposition with a new predicate or timing bound $\pi_{\mu'}$. In our example, during execution, we modify $\gls{psistl}_{alarm}$ to be Eqn. \ref{modifiedEqn}.

\begin{equation}
\begin{split}
    \gls{psistl}_{new} &= (G(alarm \Rightarrow F_{[0,15]}(\parallel \textbf{x}_t - [3,5]\parallel < 0.5 )) \\ & \wedge G_{[0,50]}(\parallel \textbf{x}_t - [1,1] \parallel > 0.5 )) \\ & \vee  G(alarm \Rightarrow (F_{[0,15]} \parallel \textbf{x}_t - [0,4]\parallel < 1))
\end{split}
\label{modifiedEqn}
\end{equation}

For $\gls{psinew}$ in our example: 

\begin{itemize}
    \item $\gls{psistl}'_1 = G_{[0,50]}(\parallel \textbf{x}_t - [1,1] > 0.5 \parallel)$ 
    \item $\gls{psistl}'_2 = G(alarm \Rightarrow (F_{[0,15]} \parallel \textbf{x}_t - [0,4]\parallel < 1))$
    \item $\gls{opset} = [\wedge, \vee]$

	\item $\pi_{\parallel \textbf{x}_t - [3,4]\parallel < 1,[0,7]} \rightarrow \pi_{\parallel \textbf{x}_t - [3,5]\parallel < 0.5,[0,15]}$
\end{itemize}

\textbf{\textit{prepSpec} (lines \ref{line:prepMod}-\ref{line:addcbf}):} We prepare the specification $\gls{psistlset} [i]$ in the same manner as the original specification $\gls{psistl}$. This includes abstracting $\gls{psistlset} [i]$ as an LTL formula and generating a B\"{u}chi automaton (line \ref{line:prepMod} of Algo. \ref{algo:modification}). From abstracting $\gls{psistlset} [i]$, the abstracted propositions $\gls{abstractedpred}'$, the set of predicate functions $\gls{setpredicatefunc}'$, and the set of Control Barrier Functions $\gls{cbf}'$ are added to their respective sets. Depending on $\gls{opset} [i] $, we update the current set of B\"{u}chi automata \gls{buchiset}. 

\textbf{\textit{intersectB} (line \ref{line:buchiIntersect}):} If the specification is added via conjunction, $\gls{opset} [i] = \wedge$,  we find the B\"{u}chi intersection of all $\gls{buchi} \in \gls{buchiset}$ and the B\"{u}chi automaton for the added specification $\gls{buchi}^{'}$ following the procedure in Sec. \ref{sec:buchiIntersect} and \cite{clarke2018model}. When making this modification, the size of the set \gls{buchiset} does not change. The intersection of the automata $\gls{buchi}$ and $\gls{buchi}'$ results in a single B\"{u}chi automaton in which an accepting run satisfies both $\gls{psistl}$ and $\gls{psistl'} [i]$.

\textbf{(lines \ref{line:addbuchi} - \ref{line:addstate}):} If the specification is added via disjunction, $\gls{opset} [i] = \vee$, then the B\"{u}chi automaton $\gls{buchi}'$ and its initial state $\gls{initialstate}'$ are added to the current lists \gls{buchiset} and \gls{potentialstates}. During execution the system will determine which $\gls{buchi} \in \gls{buchiset}$ should be followed to maximize instantaneous robustness. 

It is important to note the order of operation (parentheses first, followed by conjunction, and then disjunction)
in which new specifications $\gls{psistlset} [i]$ are added to the original specification \gls{psistl} should be followed as written in $\gls{psistl}_{new}$. For example, in our motivating example a specification is added via conjunction followed by disjunction. This results in the list of B\"{u}chi automata being $\gls{buchiset} = [\gls{buchi} \times \gls{buchi}_1, \gls{buchi}_2]$. Where $\gls{buchi}$ is the B\"{u}chi automaton for the original specification $\gls{psistl}$, $\gls{buchi}_1$ is the B\"{u}chi automaton for the specification added via conjunction $\gls{psistl}'_1$, and $\gls{buchi}_2$ is the B\"{u}chi automaton for the specification added via disjunction $\gls{psistl}'_2$. An incorrect ordering of adding $\gls{psistl}'_2$ via disjunction before $\gls{psistl}'_1$ would result in a list of B\"{u}chi automaton $\gls{buchiset} = [\gls{buchi} \times \gls{buchi}_1, \gls{buchi}_2 \times \gls{buchi}_1]$. 

\textbf{\textit{modifyProp} (line \ref{line:modifyprop}): } For all modifications of the predicate or timing bounds that are made to $\gls{psistl}$ we first determine if the proposition that is being modified exists in a transition to the forward reachable states of all $\gls{buchi} \in \gls{buchiset}$. That is, if the proposition will ever be required to be $True$ in any transition that is reachable from the current states of all $\gls{buchi} \in \gls{buchiset}$. If the proposition is in the forward reachable states, this means that the proposition may be activated in the future and we modify the predicate or timing bounds of the abstracted proposition.  If the proposition does not exist in a transition to the forward reachable states, this means that the system has already satisfied the proposition and it will not be satisfied again.  In this case we notify the user that the modification is no longer relevant for the task. 

For example, consider a specification where a robot is tasked with eventually being in a room between times 5 and 15. After the robot has successfully visited the room within the time bounds, the system will not require the proposition that abstracts the task to be $True$ again during the execution.  The user may make a modification to the specification which changes the location of the room that the robot is to eventually visit.  If the user modifies the specification after the robot has already visited the original room location,  the robot will not visit the new room location. In this case, the system recommend that the user add the new task via conjunction. 

After all modifications are made, the Event-based STL specification \gls{psistl} is replaced with the new specification $\gls{psistl}_{new}$. This is so that, in subsequent timesteps, additional new modifications may be added. 

\subsection{Determining a Specification to Satisfy (Algo. \ref{algo:Control})}
Before execution, we initialize the lists $\gls{buchiset}$ and $\gls{potentialstates}$ (lines \ref{line:buchiStates}-\ref{line:potStates}) to store the B\"{u}chi automaton and its current state for the original specification $\gls{psistl}$; we add B\"{u}chi automata and initial states for any specifications $\gls{psistl}'$ added via disjunction through online modifications (line \ref{line:modify}) .

\textbf{\textit{chooseProps} (line \ref{line:choose}):} When a specification is modified using disjunction, the system must choose which $\gls{buchi} \in \gls{buchiset}$ it should follow to satisfy the specification. We determine which B\"{u}chi automaton and specification the system should follow by evaluating the instantaneous robustness of all B\"{u}chi automata. By doing this we choose to follow and satisfy the specification which maximizes robustness for the system.  
Using the procedure from \cite{gundana2022event}, at each time step we calculate the instantaneous robustness of the shortest path to an accepting state in each $\gls{buchi} \in \gls{buchiset}$. We present two methods to determine which $\gls{buchi} \in \gls{buchiset}$ should be executed: 

\begin{enumerate}
    \item We choose the $\gls{buchi} \in \gls{buchiset}$ that has the maximum instantaneous robustness and remove all other \gls{buchi} from $\gls{buchiset}$. This results in $|\mathcal{B}| = 1$  and essentially means that the system commits to one of the formulas in the disjunction until another modification is made. 
    \item For the current timestep we choose to execute the $\gls{buchi} \in \gls{buchiset}$ that maximizes robustness  but we do not remove any other \gls{buchi}. Instead, at each timestep we re-evaluate each $\gls{buchi} \in \gls{buchiset}$ and always choose to execute the $\gls{buchi}$ with maximum instantaneous robustness. 
\end{enumerate}

The trade-off between these methods is that in the first method, we save computation time by immediately committing to one B\"{u}chi automaton, essentially satisfying one clause of the disjunction, at the risk of failing to fulfil the full specification (if, for example, the environment changes and that part of the specification can no longer be completed in time). The second method, on the other hand, may be more robust to environment changes, as it recalculates the instantaneous robustness at every time step; however, this requires more computation time and might include chattering, where the system switches between different behaviors. For both methods, if the environment is adversarial, we cannot guarantee task completion. In this paper we follow the second method. We leave further analysis regarding the best choice of the B\"{u}chi automaton  to future work.  

\section{PHYSICAL DEMONSTRATION}
We demonstrate the utility of modifying a specification at runtime through several scenarios with a mobile manipulator  picking up objects from known locations and moving them to known depots.

\subsection{Robot and Environment Model}
\begin{figure}[h!]
    \centering
    \includegraphics[width=.5\linewidth]{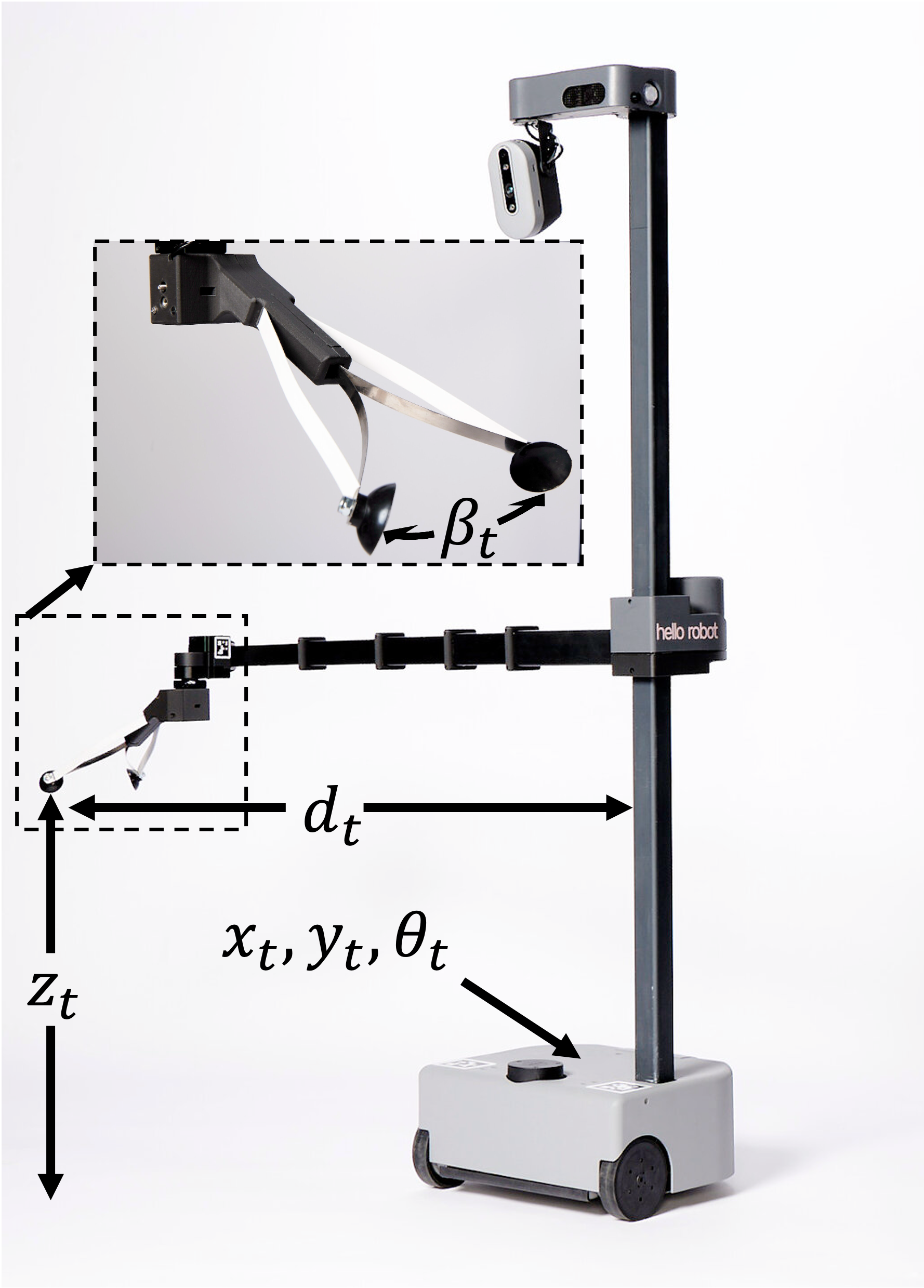}
    \caption{Hello Robot Stretch \cite{hellorobot} and its state $\textbf{x}_t = [x_t, y_t, \theta_t, d_t, z_t, \beta_t]$.
    }
    \label{stretch}
\end{figure}

We use a Hello Robot Stretch (Fig. \ref{stretch}); Eqn. (\ref{state}) represents the state of the robot. 
\begin{equation}
    \textbf{x}_t = [x_t, y_t, \theta_t, d_t, z_t, \beta_t]
    \label{state}
\end{equation}

Where $x_t$, $y_t$, and $\theta_t$ are the position and orientation of the robot base in 2D, $d_t$ is the distance from the robot origin to the tip of the gripper, $z_t$ is the height of the tip of the gripper from the ground, and $\beta_t$ represents the distance between the fingers of the gripper. 

We model the base of the robot as a holonomic robot and use feedback linearization to generate wheel velocity commands for the differential drive base. This makes the model of the system linear wrt the control. The control $\textbf{u}_t$ of the Stretch are as follows:  
\begin{equation}
    \textbf{u}_t = [v_{x,t}, v_{y,t}, \omega_t, v_{d,t}, v_{z,t}, v_{\beta,t}],
\end{equation}


where $v_{x,t}$ and $v_{y,t}$ are the velocities of the base in the x and y direction, $\omega_t$ is the rotational velocity of the base, $v_{d,t}$ is the velocity of the arm extension, $v_{z,t}$ is the velocity of the height of the arm, and $v_{\beta,t}$ is the velocity of the gripper.

We assume that the position of an object of interest $\textbf{x}_{obj,t} = [x_{obj,t},y_{obj,t},z_{obj,t}]$ and the depot where the object is deposited $\textbf{x}_{dep} = [x_{dep,t},y_{dep,t},z_{dep,t}]$ are known at all times; in practice, we use an Optitrack motion capture system to track the robot, object, and depot. 

\subsection{Original Task}
 The pick-and-place task is as follows: \textit{Consider a Stretch robot operating in a lab environment where it must pick up items and deliver them to depot areas. The Event-based STL specification is } $\gls{psistl}_{collect} = \gls{psistl}_{pick} \wedge \gls{psistl}_{align} \wedge \gls{psistl}_{extend} \wedge \gls{psistl}_{retract} \wedge \gls{psistl}_{deposit}$ where:

 \begin{itemize}[labelindent=0.5pt,itemindent=0pt,leftmargin=*]
    \item $\gls{psistl}_{pick} = G(pick \Rightarrow (F_{[0,30]}(d_{obj} < 1)))$
    \begin{itemize}
        \item Given an external event $pick$, the robot has 30 seconds to get to the location of the object.
    \end{itemize}

    \item $\gls{psistl}_{align} = G(d_{obj}<1 \Rightarrow (F_{[0,15]}( \theta_{obj} < 0.1)))$
    \begin{itemize}
        \item Whenever the robot is within 1 unit of the object, it must align its body and arm with the object within 15 seconds.  
    \end{itemize}
    
    \item $\gls{psistl}_{extend} = G((d_{obj}<1 \wedge \theta_{obj} < 0.1) \Rightarrow (F_{[0,10]}(|z_t - z_{obj,t}| < 0.05 \wedge |d_t - d_{obj,t}|< 0.05)) \wedge F_{[10,15]}(\beta_t < 1)))$
    \begin{itemize}
        \item If the robot is within 1 unit and aligned with the object, it must first extend its arm to reach the location of the object within 10 seconds. After 10 seconds but before 15 seconds, the distance between the gripper's fingers must be less than 1 (the gripper is closed). 
    \end{itemize}

    \item $\gls{psistl}_{retract} = G((\beta_t<1)\Rightarrow ((\beta_t < 1) U_{[0,25]}(d_{dep}<1 \wedge d_t < 0.2 \wedge |z_t - z_{dep}|< 0.1)))$
    \begin{itemize}
        \item If the distance between the gripper fingers is less than 1 (the robot has grasped the object), the robot must maintain this gripper distance until it eventually within 25 seconds reaches the location of the depot. Additionally the robot arm must eventually be retracted to be close to the body of the robot and the height of the gripper must eventually be within 0.1 units of the height of the depot. 
    \end{itemize}
    \item $\gls{psistl}_{deposit} = G((d_{dep}<1) \Rightarrow (F_{[0,20]}(\theta_{dep} < 0.1 \wedge |d_t-d_{dep}|<0.05) \wedge F_{[20,25]}(\beta_t > 3)))$
    \begin{itemize}
        \item Whenever the robot is within 1 unit of the depot, the robot must align with the depot and extend its arm so that the gripper reaches the depot within 20 seconds. After 20 seconds but before 25 seconds of being within 1 unit of the depot, the distance between the grippers must be greater than 3 which releases the object.
    \end{itemize}
\end{itemize}

In the specification $d_{obj} = \parallel [x_t,y_t] - [x_{obj,t},y_{obj,t}] \parallel$, $d_{dep} = \parallel [x_t,y_t] - [x_{dep,t},y_{dep,t}] \parallel$, $\theta_{obj} =|\theta_t -tan^{-1}(\frac{y_{obj,t}-y_t}{x_{obj,t}-x_t}) + \frac{\pi}{2}|$, and $\theta_{dep} =|\theta_t -tan^{-1}(\frac{y_{dep,t}-y_t}{x_{dep,t}-x_t}) + \frac{\pi}{2}|$. The angles $\theta_{obj}$ and $\theta_{dep}$ align the robot perpendicular to the object/depot such that the arm, which extends outward and does not rotate, is aligned with the object/depot. 

\subsection{Online Modifications}
\begin{figure}[h!]
    \centering
    \includegraphics[width=.7\linewidth]{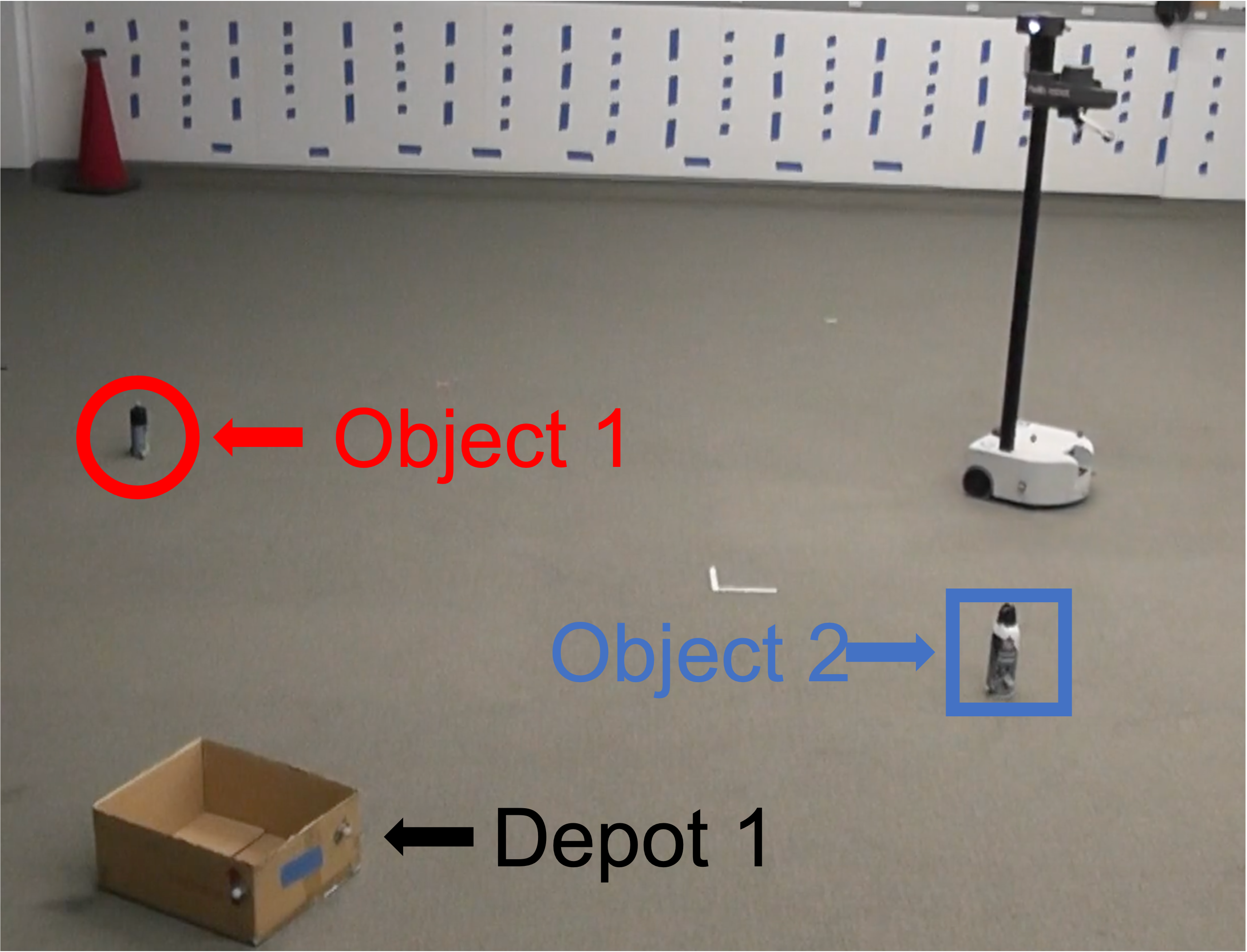}
    \caption{Physical demonstration setup. Object 1 is shown in the red circle, Object 2 (used for modification 2) is in the blue square, the depot is in the bottom left corner, and the robot is in the top right corner.}
    \label{physical}
\end{figure}

During execution, the user can make online modifications to the specification in response to pre-failure warnings or in order to change the robot's behavior. We demonstrate such  modifications below and in our accompanying video. Figure \ref{physical} shows the setup of the demonstration and one potential initial configuration of the robot, object, and depot. 

1) During execution there may not be enough time for the robot to reach the object. This may be due to disturbances while navigating or an initial robot position that is too far away from the object. In the demonstration, the system generates a pre-failure warning indicating that there is not enough time to reach the location of the object. 
Here, the user is able to modify the specification to add more time for the robot to reach the object and allowing the remainder of the specification to be satisfied. We modify $\gls{psistl}_{pick}$ to be the following: 
\begin{itemize}
    \item $\gls{psistl}_{pick} = G(pick \Rightarrow (F_{[0,45]}(d_{obj} < 1)))$
\end{itemize} 
During execution, this gives the robot 15 more seconds to reach the object without causing the specification to be violated. 

2) To demonstrate the ability to change predicate functions we modify $d_{obj}$ such that a different object that is placed in the environment is picked up by the robot. During execution we change each occurrence of $d_{obj}$ in $\gls{psistl}_{collect}$ to be $d_{obj,2}$, where $d_{obj,2} = \parallel [x_t,y_t] - [x_{obj,2,t},y_{obj,2,t}] \parallel$. Additionally, we modify the predicate $d_t<0.2$ in $\gls{psistl}_{retract}$ to be $d_t<0.05$. This results in the arm being closer to the robot's body while navigating. This aids in ensuring the object does not fall from the gripper as it navigates to the depot. 

3) In the third demonstration we add additional specifications via conjunction. The modified specification is $\gls{psistl}_{new} = \gls{psistl}_{collect} \wedge \gls{psistl}_{avoid}$. Where $\gls{psistl}_{avoid} = G_{[0,100]}(\parallel [x_t,y_t] - [cone_{x,1}, cone_{y,1}]\parallel > 0.3\  \wedge \parallel [x_t,y_t] - [cone_{x,2}, cone_{y,2}]\parallel > 0.3)$. This newly modified specification requires the robot to avoid two cones that are placed in the environment during execution. The location of the cones are known to the robot through the Optitrack system.  

4) The final demonstration modifies the specification to allow the object to be deposited in either depot 1 or depot 2. To do this we modify the specification to be $\gls{psistl}_{new} = \gls{psistl}_{collect} \vee \gls{psistl}_{collect'}$. Where $\gls{psistl}_{collect'} = \gls{psistl}_{pick} \wedge \gls{psistl}_{align} \wedge \gls{psistl}_{extend} \wedge \gls{psistl}_{retract'} \wedge \gls{psistl}_{deposit'}$. All instances of the position $d_{dep}$ in $\gls{psistl}_{retract} \wedge \gls{psistl}_{deposit}$ are replaced with $d_{dep,2}$ which represents the location of a second depot $d_{dep,2} = \parallel [x_t,y_t] - [x_{dep,t,2},y_{dep,t,2}] \parallel$. By adding $\gls{psistl}_{collect'}$ via disjunction, the system chooses to navigate and deposit the item in the depot such that robustness is maximized at each timestep. In the demonstration the locations of the depots are changed in real time and the system chooses which depot the object should be placed based on the time remaining in the specification and the location of the depots. 

\subsection{Demonstration Discussion}

In the accompanying video we show a demonstration of each modification. Here, we discuss the computation limitations and results of the modifications. In the first demonstration the robot is able to complete the task as it is written. Following work from \cite{gundana2022event} and line \ref{line:prepare} of Algo. \ref{algo:Control} the specification is pre-processed. The time it takes to prepare a specification, known as prep time, is the time required to abstract each predicate, create a B\"{u}chi automaton from the abstracted LTL formula using Spot \cite{Duret-Lutz2016}, and to parse the output from Spot into a readable format. Prep time is a one-time preprocessing time for a given specification. The total prep time for our demonstration was 3.74 seconds. During execution of the first demonstration there are no pre-failure warnings given and the average computation time to generate a control for the robot at each time step was 0.14 seconds. All demonstrations are run on a 2.3 GHz Quad-Core CPU with 8 GB of RAM.

In modification 1, the initial position of the robot was changed so that the robot did not have enough time to reach the object in the required time. During execution a pre-failure warning was given that the specification may be violated in the future and $\Psi_{pick}$ was modified to increase the time from 30 to 45 seconds. The average computation time for the execution of the specification was 0.14 seconds and the total modification time took 0.005 seconds to complete. This includes the time it takes to identify the modified timing bound and change all appropriate abstracted predicates to represent the modification. 

The total modification time to find the predicate change in the specification and modify the abstracted predicates and CBFs for modification 2 was 0.009 seconds. In these cases both the modifications for timing bound and predicate change modifications do not require the execution to be stopped or paused because no pre-failure warning was given after the modification was made. The modification time only occurs for one time step and computation times for future time steps are not impacted by the modification. 

For the third modification a new specification was added via conjunction. The total modification time for this modification includes the time to identify the new specification, abstract it as an LTL specification, create a B\"{u}chi automaton, and find a B\"{u}chi intersection with all existing B\"{u}chi automata in the system. The most computationally expensive operations are the B\"{u}chi automaton generation with Spot \cite{Duret-Lutz2016} and finding a B\"{u}chi intersection with all B\"{u}chi in $\gls{buchiset}$. For this demonstration, there was only one existing B\"{u}chi automaton. B\"{u}chi automaton are added to the system when specifications are added via disjunction. The complexity of the specification being added via conjunction greatly increases the modification time and, if the modification time is large, a pause may be needed in execution. In our demonstration the total modification time of adding the collision avoidance specifications took 0.23 seconds. In this case, this is fast enough to not require the execution to pause. For the addition of more complex specifications whose translation to a B\"{u}chi automata require significant computation time, a pause in execution may be required. 

The final modification adds a new specification $\gls{psistl}_{collect'}$ via disjunction. This specification has the same complexity as the original specification $\gls{psistl}_{collect}$. Because only predicates are changed between the two specifications the abstracted LTL formulas are identical and they result in the same B\"{u}chi automaton. The total modification time to add a specification via conjunction includes the time to identify the new specification being added, abstract it as an LTL formula, and generate a B\"{u}chi automaton. In this example, the modification time of $\sim$3.75 seconds was similar to the prep time for the original specification $\gls{psistl}_{collect}$ because of the identical B\"{u}chi automata. In some cases, this longer computation time may require an execution pause in order to satisfy the specification. When adding specifications via disjunction, the system is given the choice of which specification to satisfy and it chooses the one that maximizes instantaneous robustness. As a result of searching more B\"{u}chi automata, the average computation time rises from 0.14 seconds to 0.16 seconds.

\section{Conclusions}
In this paper we formally define a modification grammar for Event-based STL specifications. We present an automated control synthesis framework to satisfy, online, modifications of Event-based STL specifications, and provide feedback for modifications that may no longer be relevant based the timing of the modifications. Through our demonstrations we show that online modifications of predicates and timing bounds are computationally efficient and do not require a pause in execution. 

Currently, the computation time of making online modifications of adding specifications via conjunctions or disjunction are limited by the time it takes to generate a B\"{u}chi automaton which increases with the complexity of a specification. In future work we will determine the feasibility of reducing this computation time by pre-computing complex B\"{u}chi automata for known modifications or exploring other potential methods to reduce the complexity of transforming an LTL formula to an automaton. 
Additionally, we will explore methods to efficiently compute the robustness of an full trace to an accepting state in the B\"{u}chi automaton rather than a single transition. Allowing for a robust solution over the entire trajectory for the system. 



\bibliographystyle{ieeetr}
\bibliography{citations.bib}

\end{document}